\documentclass[11pt,a4paper]{article}
\usepackage[hyperref]{emnlp2018}
\usepackage{times}
\usepackage{latexsym}

\usepackage{url}

\usepackage{graphicx}
\usepackage{amsmath}
\usepackage{bm, upgreek}
\usepackage{amssymb}
\usepackage{array, booktabs}
\usepackage{tabularx}
\usepackage{multirow}
\usepackage{mathtools}
\usepackage{arydshln}
\usepackage{tabularx}

\usepackage{cleveref}
\crefformat{section}{\S#2#1#3} 
\crefformat{subsection}{\S#2#1#3}
\crefformat{subsubsection}{\S#2#1#3}

\aclfinalcopy 

\newcommand{\sect}[1]{Section \ref{#1}}

\newcommand{\squad}{SQuAD}
\newcommand{\triviaqa}{TriviaQA}

\DeclareMathOperator*{\argmax}{argmax}

\newcommand{\complexqa}{F_{\theta}}
\newcommand{\paramp}{\mathrm{Pr}}

\newcommand{\encq}{G_{\theta}}

\newcommand{\encd}{H_{\theta}}

\newcommand{\QA}{{Phrase-Indexed Question Answering}}
\newcommand{\qashort}{{\sc PIQA}}

\newcommand{\web}{\url{nlp.cs.washington.edu/piqa}}

\title{Phrase-Indexed Question Answering: \\A New Challenge for Scalable Document Comprehension}

\author{Minjoon Seo$^{2,3}$\thanks{Most work done during internship with Google AI.}\quad Tom Kwiatkowski$^1$\quad Ankur P. Parikh$^1$ \\ {\bf Ali Farhadi}$^{2,4,5}$\quad{\bf Hannaneh Hajishirzi}$^2$ \\
  Google AI Language$^1$\quad University of Washington$^2$\quad Clova AI, NAVER$^3$\\Allen Institute for AI$^4$\quad XNOR.AI$^5$ \\
  {\tt \{minjoon,ali,hannaneh\}@cs.uw.edu}\\{\tt \{tomkwiat,aparikh\}@google.com}}

\begin{document}
\maketitle

  
  
  
  

\begin{abstract}
We formalize a new modular variant of current question answering tasks by enforcing complete independence of the document encoder from the question encoder. This formulation addresses a key challenge in machine comprehension by requiring a standalone representation of the document discourse. It additionally leads to a significant scalability advantage since the encoding of the answer candidate {\it phrases} in the document can be pre-computed and {\it indexed} offline for efficient retrieval.  
We experiment with baseline models for the new task, which achieve a reasonable accuracy but significantly underperform unconstrained QA models. We invite the QA research community to engage in \QA~(\qashort, {\it pika}) for closing the gap. The leaderboard is at: \web

\end{abstract}

\section{Introduction}
Extractive question answering (QA) is the task of selecting an answer phrase (span) to a question  given an evidence document.
Due to the easiness of evaluation (compared to generative QA) and the fine-grainess of the answer (compared to sentence-level QA),
it has become one of the most popular QA tasks, driven by
massive new datasets such as \squad~\cite{Rajpurkar2016SQuAD10} and \triviaqa~\cite{joshi2017triviaqa}.
Current QA models heavily rely on  explicitly learning the interaction between the evidence document and the question using neural attention mechanisms ~\citep[\textit{inter alia}]{wang2016machine,xiong2016dynamic,seo2016bidirectional,lee2016learning}, in which the model is fully aware of the question before or as it reads the document.  As a result, despite significant advances, they have not led to the standalone representation of document discourse which is nevertheless a key goal of research in reading comprehension. Furthermore, QA models that condition the document representation on a question have the practical scalability downside that the entire model should be re-applied on the same document for every question. 

\begin{figure}[t]
    \centering
    \includegraphics[width=.9\linewidth]{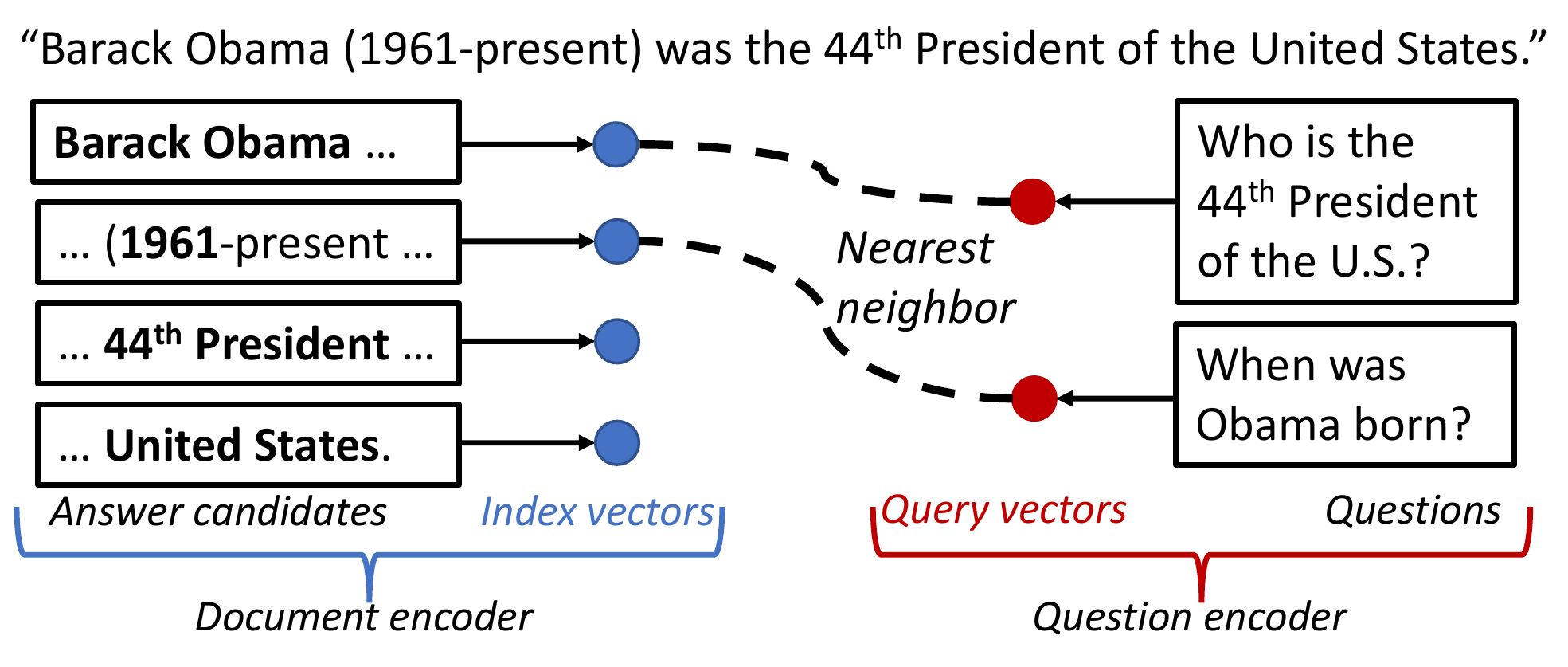}
    \vspace{-.3cm}
    \caption{\qashort\ task for a short context sentence.}
    \label{fig:teaser}
    \vspace{-1mm}
\end{figure}

In this paper, we formalize a modular variant of the QA task, Phrase Indexed Question Answering (\qashort), that enforces complete independence between document encoder and question encoder (Figure~\ref{fig:teaser}).  In \qashort, all documents are processed independently of any question to generate {\it phrase index vectors} (blue nodes in the figure) for each answer candidate (left boxes in the figure).  Similarly, the questions are  independently mapped to {\it query vectors} (red nodes in figure).  Then, at inference time, the answer is obtained by retrieving the nearest indexed phrase vector to the query vector.
Hence the algorithms aimed at tackling  \qashort\  have the inherent benefit of modularity and scalability compared to current QA systems. 

The task setup is analogous to how documents or sentences are retrieved in modern search engines via similarity search algorithms~\cite{shrivastava2014improved}. Nevertheless, there is a key distinction that search engines index each \emph{document} by its \emph{content}, while \qashort\ requires one to index each \emph{phrase} in documents by its \emph{context}.

We formally define the \qashort\ problem and provide baseline models for the new task. Our experiments show that  the constraint introduced by \qashort\ leads to meaningful standalone document representations and practical scalability advantage, demonstrating the significance of the new task. 
Moreover, there is still a large gap between the baselines and the unconstrained state of the art, showing that the task is yet far from being solved. We have set up a leaderboard\footnote{\web}for \qashort\ challenge and invite the research community to participate. We currently support SQuAD and plan to expand to other datasets as well.

\section{Related Work}\label{sec:related}
\vspace{-.2cm}
\paragraph{Reading comprehension.}
Massive reading comprehension question answering datasets~\cite{hermann2015teaching, hill2015goldilocks, dhingra2017quasar, dunn2017searchqa} have driven a large number of successful neural approaches~\citep[\textit{inter alia}]{kadlec2016text,hu2017reinforced}.  
~\citet{choi2017coarse,chen2017reading,clark2017simple,min2018efficient} tackled large-scale QA by using a fast, coarse model (e.g. TF-IDF) to retrieve few documents or sentences and then using a slower, accurate model to obtain the answer.
\citet{salant2017contextualized} proposed to minimize (but not prohibit) the influence of question when modeling the document.
Similarly to ours, \citet{lee2016learning} proposed to explicitly learn the representation for each answer candidate (phrase) in the document, but it was conditioned (dependent) on the question.



\vspace{-.2cm}
\paragraph{Sentence retrieval.} A closely related task to ours is that of retrieving a sentence/paragraph in a corpus that answers the question~\citep{tay2017hyperqa}. A comprehensive survey for neural approaches in information retrieval literature is discussed in~\citet{mitra2017neural}. 
We note that our problem is focused on phrasal answer extraction, which presents a unique challenge over sentence retrieval---the need for \textit{context-based} representation as opposed to the \textit{content-based} representation in the sentence-retrieval literature.


\vspace{-.2cm}
\paragraph{Language representation.} Recently there has been a growing interest in developing natural language representations that can be transferred across tasks~\citep[\textit{inter alia}]{vendrov2015order,wieting2015towards, conneau2017supervised}.
In particular, SNLI~\cite{SNLI} and MultiNLI~\cite{williams2017broad} encourage architectures that first encode the hypothesis and the premise independently before a comparator neural network is applied.
Our proposed problem shares similar traits but has a stronger constraint that only inner product comparison is allowed and one needs to model phrases instead of complete sentences.






\paragraph{Memory networks.} Each phrase-vector is analogous to a single memory slot, where the vector is the key and the phrase is the value, and the question vector is the query for accessing the memory. Hence, \qashort\ can be considered as an effort to formulate extractive question answering as the task of memory augmentation (construction)~\cite{memory, sukhbaatar2015end} from unstructured knowledge source (text).

\section{\QA}\label{sec:problem}
Extractive question answering is the task of obtaining the answer $\hat a$ to a question $Q = \{q_1 \dots q_n\}$ given an evidence document $D = \{ d_1 \dots d_m \}$, where the answer $\hat a = (s, e)$ indicates the start and end of a span in the document.
The task is often formulated as learning the probabilistic distribution of the answer given the question and the document. 
In existing literature (\sect{sec:related}), the distribution is mainly featurized by $\paramp(a|Q,D) \propto \exp(\complexqa(Q, D, a))$ where $\complexqa$ could be any real-valued scoring function parameterized by $\theta$.
Once $\theta$ is learned, the prediction $\hat{a}$ is obtained by
\begin{equation}
\hat a = \argmax_{a} \complexqa(Q, D, a).
\end{equation}
So far, most competitive designs of $\complexqa(Q,D,a)$ make use of attention connections between the words in $Q$ and $D$.
As a result, these models cannot yield a query independent representation of the document $D$. It is subsequently not possible to independently assess the document understanding capability of the model. Furthermore, $\complexqa(Q, D, a)$ needs to be re-computed for the entire document for every new question. We believe that this inefficiency  precludes all current models as the candidates for end-to-end QA systems.

We propose a new task---\QA\ (\qashort)---that addresses these issues. 
We enforce the \emph{decomposability} of $\complexqa$ into two exclusive functions $\encq(Q), \encd(D, a) \in \mathbb{R}^k$. 
The answer distribution is then modeled by $\paramp(a|Q,D) \propto \exp(\encq(Q) \bullet \encd(D, a))$, where $\bullet$ is the inner product. The prediction is obtained by 
\vspace{-0.5mm}
\begin{equation}
  \hat a = \argmax_{a} \encq(Q) \bullet \encd(D, a).
  \label{eqn:form2}
\end{equation}
In this setting, the document encoder $\encd$ learns models the document independently of the question.
Successful question answering models that follow the structure of \qashort\ will have two important advantages over current QA models: full document comprehension and scalablity. 

\paragraph{Full document comprehension.}  Language understanding ability is widely associated with learning a good standalone representation of text (or its components such as phrases) independent of the end task~\cite{SNLI}. Under \qashort\ constraints, the document encoder $\encd$ learns the representation of the answer candidate phrases $a$ in the document $D$ independent of the question. In order to correctly answer questions, these  phrase representations (index vectors) need to correctly encode their meaning with respect to their context. 
Therefore,  \qashort\ constraint enforces evaluating research in document comprehension and phrase representation learning.


\paragraph{Scalability.} Models that adhere to the \qashort\ constraint only need to be run once for each document, regardless of the number of questions asked. 
To answer a question, the model then just needs to encode the question and compare it to each of the answer candidates via the inner product in Equation~\ref{eqn:form2}.
Implemented naively, computing a single inner product for each answer candidate is more efficient than building a new document encoding; after the documents are pre-encoded, Equation~\ref{eqn:form2} is $O(k)$ time per word where $k$ is the vector size (most neural models require $O(k^2)$ per word for matrix multiplications).

More importantly, 
\qashort\ also permits an approximate solution in sublinear time using asymmetric locality-sensitive hashing (aLSH)~\cite{shrivastava2014asymmetric, shrivastava2014improved},
through which Equation~\ref{eqn:form2} can be approximated for $N$ answer candidates with $O(k N^\rho  \log N)$ time, where $\rho < 1$ is a function of the approximation factor and the properties of the hash functions.
We argue that this type of approach will be essential for the development of real world QA systems, where the number of potential answers $N$ is extremely large.

\section{Baseline Models}\label{sec:baselines}
We introduce several baselines for \qashort\ that are motivated by related literature.

For all (neural) baselines, we represent the words in $D$ and $Q$ with one of three embedding mechanisms:
CharCNN~\cite{kim2014convolutional} + GloVe~\cite{pennington2014glove}, and ELMo~\cite{peters2018deep}.
We follow the majority of the related literature and apply bidirectional LSTMs~\cite{Hochreiter:97} to these embeddings to build the context-aware representations of the document ${\bf D} = \{{\bf d}_1 \dots {\bf d}_m\}$ and question ${\bf Q} = \{{\bf q}_1\dots {\bf q}_n\}$, where the forward \& backward LSTM outputs are concatenated to get a single word representation, i.e. ${\bf d}_i, {\bf q}_i \in \mathbb{R}^{2k}$ where $k$ is the hidden state size of LSTMs.

\qashort\ disallows cross-attention between document and question.
However, we can still benefit from self-attention, which has become crucial for machine translation~\cite{vaswani2017attention} and QA~\cite{huang2017fusionnet, wei2018fast}.
In all of our baselines, each variable-length question is collapsed into a fixed length vector via the sum 
 ${\bf q}^\text{SA} = \sum_i u_i {\bf q}_i$ where ${\bf u} = \{u_1\dots u_n\}$ is a vector containing a single weight for each word in the question. 
Similarly, we experiment with document side self attention to represent each document word ${\bf d}_j$ as a weighted sum of itself and all neighboring words ${\bf d}^\text{SA}_j = \sum_i h^j_i {\bf d}_j$.
The weight vectors $\bf u$ and ${\bf h}^j$ are calculated as
\begin{align}
    {\bf u} &= \mathrm{softmax}_i({\bf w}^\top {\bf q}_i) \notag \\ 
    {\bf h}^j &= \mathrm{softmax}_i(R_{\theta}({\bf D}, j)^\top K_{\theta}({\bf D}, i)) \notag
\end{align}
where $R_\theta$, and $K_\theta$ are trainable neural networks with the same ouptut size, and ${\bf w} \in \mathbb{R}^{2k}$ is a trainable weight vector.
We use independent BiLSTMs with hidden state size $k$ (i.e. the output size is $2k$) to model both $R_\theta$ and $K_\theta$. That is, $R_\theta({\bf D}, j)$ is the $j$-th output of BiLSTM on top of ${\bf D}$, and we similarly define $K_\theta$ with unshared parameters.

For all (neural) baselines, the question is represented using the concatenation of two copies of ${\bf q}^\text{SA}$, one that should have high inner product with the vector for the answer's start span and another that should have high inner product with the vector for the answer's end. Thus, Equation~\ref{eqn:form2}'s $\encq(Q)=[{\bf q}^\text{SA}_{s}, {\bf q}^\text{SA}_{e}]$ where the subscripts $s$ (start) and $e$ (end) imply that different sets of parameters were used.
Now we define several baselines.

\paragraph{LSTM baseline.}
An answer candidate $a = (s,e)$ is represented using the LSTM outputs at its endpoints: from Equation~\ref{eqn:form2}, $\encd(D,(s,e)) = [{\bf d}_s, {\bf d}_e] \in \mathbb{R}^{4k}$ and $\encq(Q)=[{\bf q}^\text{SA}_{s}, {\bf q}^\text{SA}_{e}] \in \mathbb{R}^{4k}$.

\paragraph{LSTM+SA baseline.}
The LSTM outputs are augmented with the endpoint representations that come out of the document's self-attention (SA):
$\encd(D,(s,e)) = [{\bf d}_s, {\bf d}^\text{SA}_s, {\bf d}_e, {\bf d}^\text{SA}_e] \in \mathbb{R}^{8k}$ and $\encq(Q)=[{\bf q}^\text{SA}_{s1},{\bf q}^\text{SA}_{s2}, {\bf q}^\text{SA}_{e1},{\bf q}^\text{SA}_{e2}] \in \mathbb{R}^{8k}$.

\paragraph{TF-IDF.} We lastly include a purely TF-IDF-based model, where each answer candidate phrase is associated with a bag of neighbor words within a distance of 7. Then the BOW vector is normalized via TF-IDF and indexed. When the query comes in, its TF-IDF vector is queried on the indexed phrases to yield the answer.

\paragraph{}For training the (neural) models, we minimize the negative log probability of getting the correct answer: the loss function for each example $(D, Q, a^*)$ is $L(\theta) = -\log \paramp(a^*|D,Q)$ where $a^*$ is the correct answer.

\section{Experiments}\label{sec:exp}
\begin{table}
    \centering
    \resizebox{\columnwidth}{!}{%
    \begin{tabular}{clcc}
        \toprule
        Constraint & Model & F1 (\%) & EM (\%) \\
        \midrule
        \multirow{ 5}{*}{PI} & TF-IDF & 15.0 & 3.9\\
         \cline{2-4}
         & LSTM & 57.2 & 46.8\\
         & LSTM+SA & 59.8 & 49.0\\
         \cline{2-4}
         & LSTM+ELMo & 60.9 & 50.9\\
         & LSTM+SA+ELMo & 62.7 & 52.7\\

		\midrule
        \multirow{2}{*}{None}  & \citet{Rajpurkar2016SQuAD10} & 51.0 & 40.0\\ 
         & \citet{wei2018fast} & 89.3 & 82.5\\
        \bottomrule 
    \end{tabular}
    }
    \caption{Performance on \squad\ dev set with the \qashort\ constraint (top), and without the constraint (bottom). See Section~\ref{sec:baselines} for the description of the terms.}
    \label{tab:squad}
\end{table}

We impose the independence restrictions from \qashort\ on the Stanford Question Answering Dataset\footnote{PIQA paradigm can be also extended to other extractive QA datasets.}. 
We only consider answer spans with  length $\leq 7$. We use the hidden state size ($k$) of $128$, which results in a 512D ($4k$) and 1024D ($8k$) vector for each phrase in LSTM and LSTM+SA, respectively. The default embedding model is CharCNN concatenated with 200D GloVe, with an option to append ELMo vectors following the same setup for SQuAD experiments discussed in~\citet{peters2018deep}. We use a batch size of $64$ and train for $20$ epochs with the default Adam optimizer~\cite{adam}, and take the best model on the validation set during training.

\paragraph{Results.}
Table~\ref{tab:squad} shows the results for the \qashort\ baselines (top) and the unconstrained state of the art (bottom). 
First, the TF-IDF model performs poorly, which signifies the limitations of traditional document retrieval models for the task.
Second, we note that the addition of self-attention makes a significant impact on results, improving F1 by 2.6\%. 
Next, we see that adding ELMo gives 3.7\% and 2.9\% improvement on F1 for LSTM and LSTM+SA models, respectively.
Lastly, the best \qashort\ baseline model is 11.7\% higher than the first (unconstrained) baseline model~\citep{Rajpurkar2016SQuAD10} and 26.6\% lower than the state of the art~\citep{wei2018fast}.
This gives us a reasonable starting point of the new task and a significant gap to close for future work.

\begin{table}
    \centering
    \resizebox{\columnwidth}{!}{%
    \begin{tabular}{l p{0.45\textwidth}}
        \toprule
        - According to the {\bf American Library Association}, this makes\dots \\\cline{2-2}
        - \dots tasked with drafing a {\bf European Charter of Human Rights},\dots\\
        \midrule
        - {\bf The LM engines} were successfully test-fired and restarted, \dots. \\\cline{2-2}
        - {\bf Steam turbines} were extensively applied\dots\\
        \midrule
        - \dots primarily accomplished through the {\bf ductile stretching and thinning}.\\\cline{2-2}
        - \dots directly derived from {\bf the homogeneity or symmetry of space}\dots\\
        \bottomrule
    \end{tabular}%
    }
    \caption{Most similar phrase pairs from disjoint sets of documents. Bold print is the phrase, and non-bold is its context.}
    \label{tab:phrase}
\end{table}

\paragraph{Phrase representations.}
Since \qashort\ models encode all answer candidates into the same space, we expect similar answer candidates to have high inner products with one another.
Table~\ref{tab:phrase} shows pairs of answer candidates that come from different documents in SQuAD, but that have similar encodings (high inner product).
We observe that phrase representations learned through the \qashort\ task capture different interesting characteristics of the phrases.
In all three rows, we can see that the phrase pairs seem to fit into natural categories: national, or multi-national organizational constructs; mechanical engines; and mechanical properties, respectively. 
This suggests that the model has learned interesting typing information above the word level. 
The second and third rows also indicate that the model has learned a rich representation of context. 
This is particularly obvious in the third row where the two phrases are lexically dissimilar, but preceded by the similar contexts {\it `primarily accomplished through'} and {\it `directly derived from'}.
We believe that this analysis, while not complete, points toward exciting future lines of work in learning highly contextualized phrase representations through question answering.


\paragraph{Scalability.}
\qashort~can also gain massive execution time speedups once the documents are pre-encoded: in our simple benchmark on a consumer-grade CPU and NumPy (for LSTM+SA model, 1024D vectors), one can easily perform exact search over 1 million document words per second.
BiDAF~\cite{seo2016bidirectional}, an open-sourced and relatively light QA model reaching 77.5\% F1 (66.5\% EM), can process less than 1k document words per second with an equivalent computing power (after pre-encoding the document as much as possible), which is more than 1,000x slower.\footnote{The difference will be even higher with a dedicated similarity search package such as Faiss~\cite{faiss} or approximate search (Section~\ref{sec:problem}).}

\begin{figure}
    \centering
    \includegraphics[width=\linewidth]{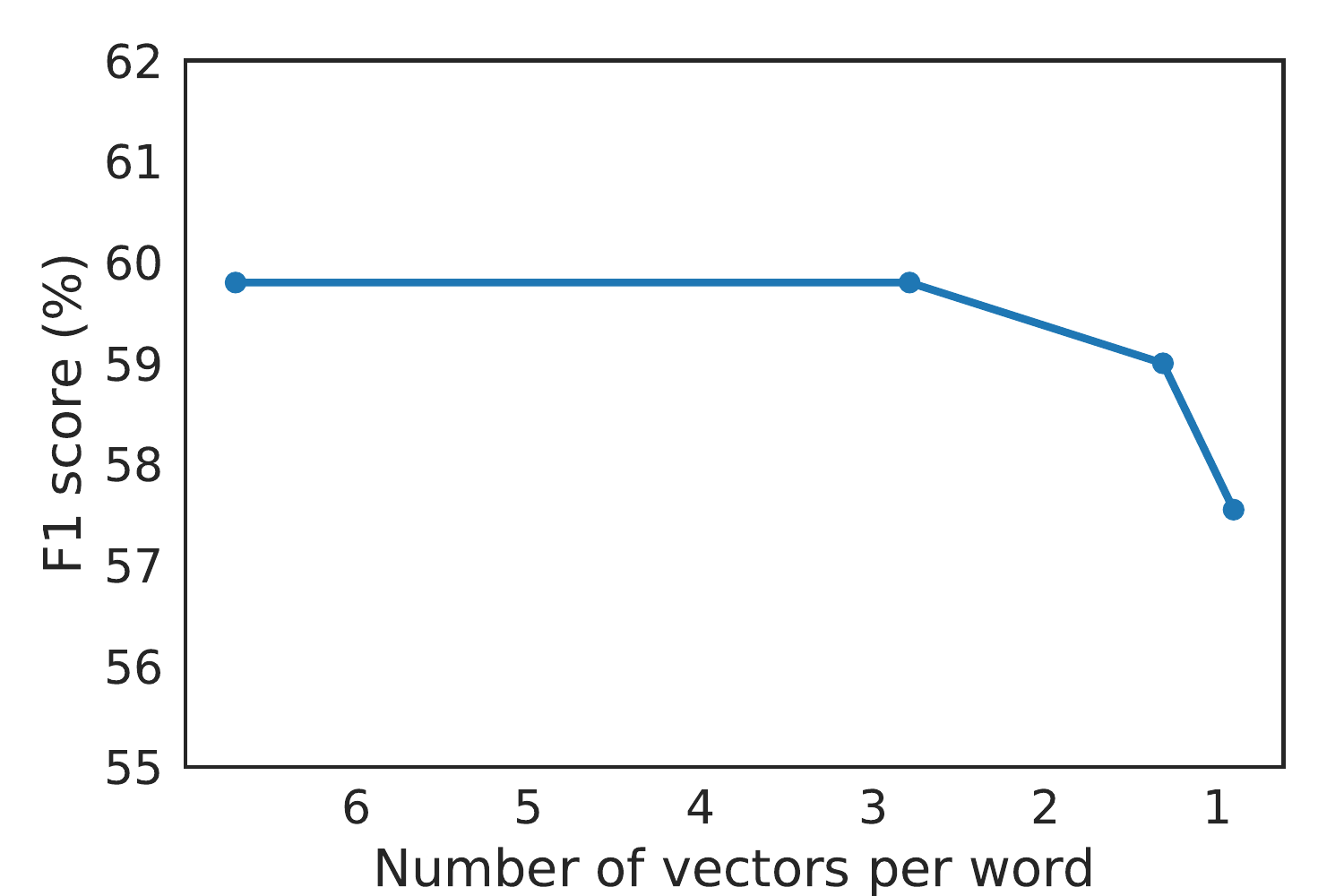}
    \caption{F1 score versus number of vectors per word for LSTM+SA. Answer candidates have been filtered with varying threshold on an independent classifier learned on the candidate representations.}
    \label{fig:compression}
\end{figure}

It is also important to consider the memory cost for storing a vector representation of each of the answer candidates.
We train an independent single-layer perceptron classifier that predicts whether the phrase encoding is likely to be a good one.
By varying a threshold on the score assigned by this classifier, we can filter answer candidates prior to storage.
Figure~\ref{fig:compression} illustrates the trade-off between accuracy and memory (measured in mean number of vectors per document word) resulting from this filtering procedure for the LSTM+SA model.
We observe that 1.3 vectors (candidates) per word on average reaches $> 98\%$ of the model's F1 accuracy. This is equivalent to 5.2 KB per word with 1024D (4 KB) float vectors, or around 15 TB for the entire English Wikipedia (3 billion words).
Future work will also involve creating a better classifier (i.e. improving the trade-off curve in Figure~\ref{fig:compression}) for determining which phrase vectors to store.

\section{Conclusion and Future Work}\label{sec:conc}
We introduced \QA\ (\qashort), a new variant of the extractive question answering task that requires documents and question encoded completely independently and that they only interact each other via inner product.
We argued that building a question-agnostic document encoder for question answering should be an important consideration for those in the QA community with the research goal of learning a model that reads and comprehends documents.
Furthermore, the imposed constraint of the task implies a sublinear scalability benefit. 
Given that SQuAD models have recently outperformed humans,
\qashort\ formulation motivates a new challenge for which we hope that the community's effort gradually closes the gap between our constrained baselines and the unconstrained models.



\subsection*{Acknowledgments}
This research was supported by ONR (N00014-18-1-2826), NSF (IIS 1616112), Allen Distinguished Investigator Award, and gifts from Google, Allen Institute for AI, Amazon, and Bloomberg. We  thank the anonymous reviewers for their helpful comments.

\bibliography{00-main}
\bibliographystyle{acl_natbib_nourl}

\end{document}